\documentclass[a4paper,fleqn]{caswenjia}
\usepackage[numbers]{natbib}
\usepackage{geometry}
\usepackage{amsmath}
\usepackage{amsfonts}
\usepackage{amssymb}
\usepackage{bbm}
\usepackage{hyperref}

\usepackage{caption}

\usepackage{amsmath}
\usepackage{url}
\usepackage{threeparttable}
\usepackage[normalem]{ulem}
\usepackage{color}
\usepackage[inline]{enumitem}

\usepackage{xcolor}
\usepackage{colortbl}

\definecolor{lightgreen}{rgb}{0, 0.69, 0.31}
\definecolor{lightblue}{rgb}{0, 0.44, 0.75}
\definecolor{brightblue}{rgb}{0, 0.69, 0.94}
\definecolor{deepred}{rgb}{0.75, 0, 0}

\def\tsc#1{\csdef{#1}{\textsc{\lowercase{#1}}\xspace}}
\tsc{WGM}
\tsc{QE}
\tsc{EP}
\tsc{PMS}
\tsc{BEC}
\tsc{DE}
\begin{document}
\begin{sloppypar}

\let\WriteBookmarks\relax
\def\floatpagepagefraction{1}
\def\textpagefraction{.001}
\shorttitle{Text-Guided Coarse-to-Fine Fusion Network for Robust Remote Sensing Visual Question Answering}
\shortauthors{Zhicheng Zhao et~al.}

\title [mode = title]{Text-Guided Coarse-to-Fine Fusion Network for Robust Remote Sensing Visual Question Answering}                      
% \tnotemark[1,2]

% \tnotetext[1]{This document is the results of the research
%   project funded by the National Science Foundation.}

% \tnotetext[2]{The second title footnote which is a longer text matter
%   to fill through the whole text width and overflow into
%   another line in the footnotes area of the first page.}
% \tnotemark[1,2]

% \tnotetext[1]{This paper was supported in part by the National Key Research and Development Program of China under Grant 2021YFB2900200, National Natural Science Foundation of China under No. 61925101, and 61831002, the Beijing Municipal Science and Technology Project NO. Z211100004421017.}

\author[1,2,3]{Zhicheng Zhao}
\address[1]{Anhui Provincial Key Laboratory of Multimodal Cognitive Computation, Anhui University, Hefei 230601, China}
\author[1,2,3]{Changfu Zhou}
\author[1,2,3]{Yu Zhang}
\author[1,2,3]{Chenglong Li}
\cormark[1]
\ead{lcl1314@foxmail.com}
\author[5]{Xiaoliang Ma}
\author[1,4]{Jin Tang}

\address[2]{
Information Materials and Intelligent Sensing Laboratory of Anhui Province, Anhui University, Hefei, China}

%\address[2]{Information Materials and Intelligent Sensing Laboratory of Anhui Province, China}
% \address[3]{The BRain and Artificial INtelligence Lab (BRAIN LAB), School of Automation, Northwestern Polytechnical University, Xi’an, 710072, China}
\address[3]{School of Artificial Intelligence, Anhui University, Hefei, China}
\address[4]{School of Computer Science and Technology, Anhui University, Hefei, China}
\address[5]{ GEOVIS Earth Technology Co., Ltd., Hefei, 230088, China}
\cortext[cor1]{Corresponding author}
\begin{abstract}
Remote Sensing Visual Question Answering (RSVQA) has gained significant research interest. However, current RSVQA methods are limited by the imaging mechanisms of optical sensors, particularly under challenging conditions such as cloud-covered and low-light scenarios. Given the all-time and all-weather imaging capabilities of Synthetic Aperture Radar (SAR), it is crucial to investigate the integration of optical-SAR images to improve RSVQA performance. In this work, we propose a Text-guided Coarse-to-Fine Fusion Network (TGFNet), which leverages the semantic relationships between question text and multi-source images to guide the network toward complementary fusion at the feature level. Specifically, we develop a Text-guided Coarse-to-Fine Attention Refinement (CFAR) module to focus on key areas related to the question in complex remote sensing images. This module progressively directs attention from broad areas to finer details through key region routing, enhancing the model's ability to focus on relevant regions.
Furthermore, we propose an Adaptive Multi-Expert Fusion (AMEF) module that dynamically integrates different experts, enabling the adaptive fusion of optical and SAR features. In addition, we create the first large-scale benchmark dataset for evaluating optical-SAR RSVQA methods, comprising 6,008 well-aligned optical-SAR image pairs and 1,036,694 well-labeled question-answer pairs across 16 diverse question types, including complex relational reasoning questions. Extensive experiments on the proposed dataset demonstrate that our TGFNet effectively integrates complementary information between optical and SAR images, significantly improving the model's performance in challenging scenarios. The dataset is available at: \url{https://github.com/mmic-lcl/}.
\end{abstract}

\let\printorcid\relax
\maketitle

\section{Introduction}

\begin{figure}
    \centering
    \includegraphics[width=8.8cm]{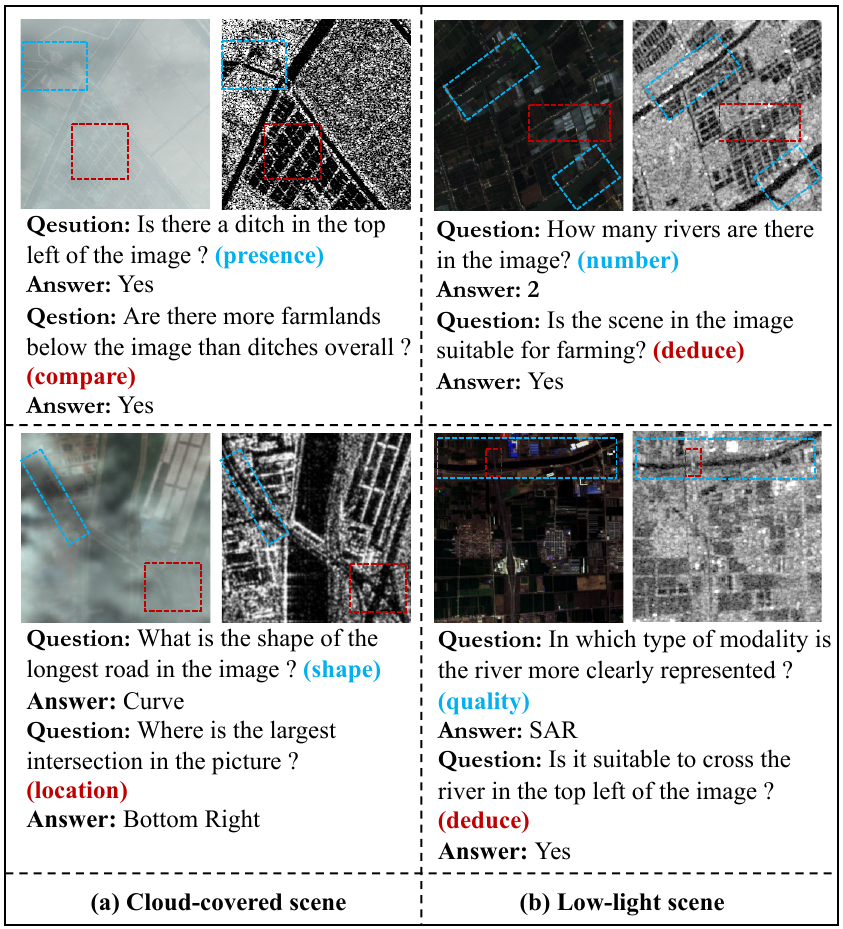}
\caption{RSVQA applications are explored in (a) cloud-covered and (b) low-light scenarios, using optical-SAR image pairs with corresponding question-answer examples. Optical images degrade significantly in these conditions, while SAR images remain robust, highlighting SAR's potential to enhance RSVQA performance in challenging environments. \textcolor{brightblue}{Blue} and \textcolor{deepred}{red} indicate regions associated with different questions within the same image pair. The text in parentheses denotes the type of question.} 
\label{fig1}
\end{figure}

{V}{isual} question answering (VQA)~\cite{VQA} has emerged as a promising research direction aimed at understanding image scenes and reasoning out answers by integrating the language information from the questions.
%understanding and inferring the rich geospatial information and complex relationships present in remote sensing imagery. 
RSVQA extends the capabilities of traditional remote sensing analysis by enabling intuitive, question-driven exploration of satellite and aerial imagery. This approach has broad applications in various domains, including land use and land cover analysis~\cite{helber2019eurosat}, urban planning~\cite{liu2020three}, environmental monitoring~\cite{li2020review}, and disaster management~\cite{harb2017remote}. By leveraging the power of VQA techniques, RSVQA allows users to obtain valuable insights from remote sensing data through interactive question-answering interfaces. This not only enhances the accessibility and usability of remote sensing technology but also facilitates more efficient and targeted analysis of large-scale geospatial datasets. \par
%RSVQA extends the capabilities of traditional remote sensing analysis by enabling intuitive, question-driven exploration of satellite and aerial imagery. This approach has broad applications in various domains, including land use and land cover analysis, urban planning, environmental monitoring, and disaster management~\cite{RSVQA}. By leveraging the power of VQA techniques, RSVQA allows users to obtain valuable insights from remote sensing data through interactive question-answering interfaces. This not only enhances the accessibility and usability of remote sensing technology but also facilitates more efficient and targeted analysis of large-scale geospatial datasets. \par

\begin{figure*}
    \centering
    \includegraphics[width=.94\textwidth]{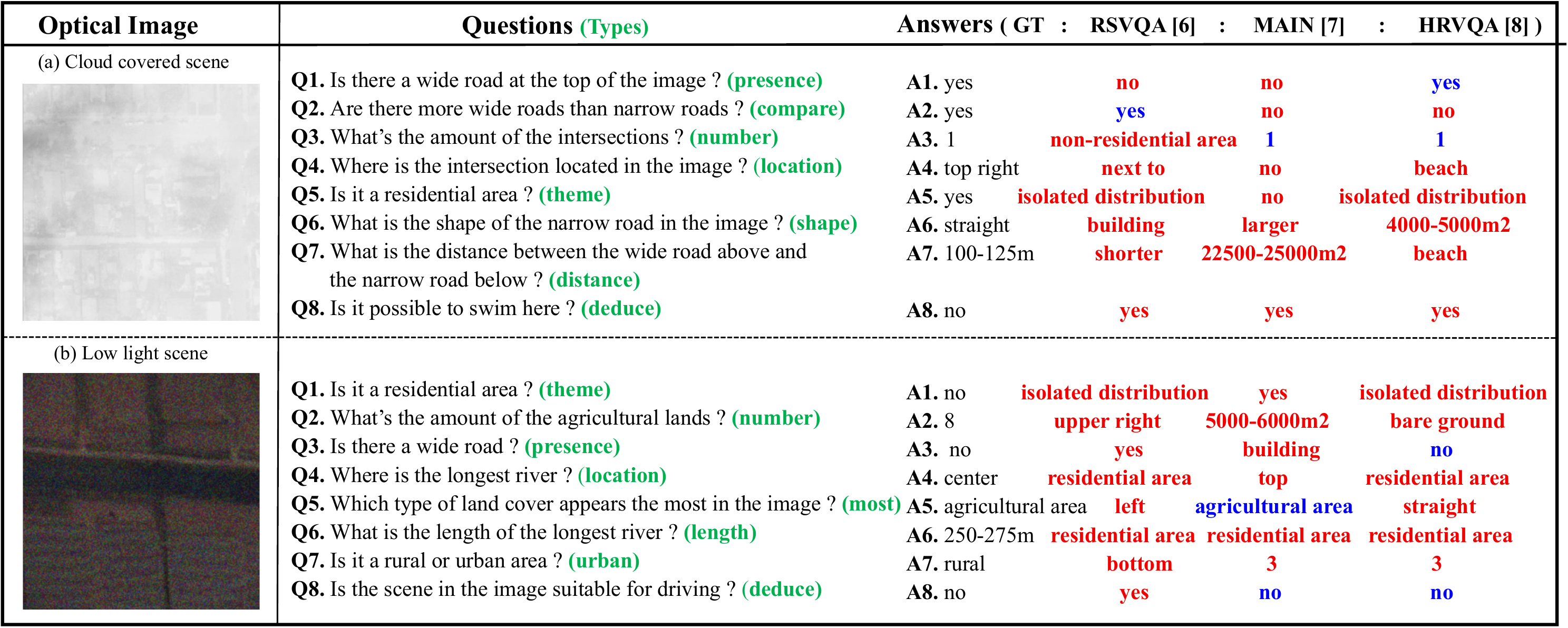}
    \caption{
    Examples of state-of-the-art (SOTA) RSVQA models~\cite{RSVQA,rsivqa, HRVQA} are evaluated in (a) cloud-covered and (b) low-light scenarios, relying solely on optical images. Question types are highlighted in \textcolor{lightgreen}{green}, correct answers are indicated in \textcolor{blue}{blue}, and incorrect answers in \textcolor{red}{red}. 
    }
    \label{fig2}
\end{figure*} 

%However, despite its potential, RSVQA faces significant challenges unique to remote sensing imagery. Remote sensing images captured from satellite platforms are often subject to conditions not typically encountered in natural scene imagery, presenting obstacles to effective analysis. As illustrated in Fig.\ref{fig1}, these adverse factors significantly hinder the performance of existing RSVQA models~\cite{RSVQA,prompt-RSVQA,rsivqa}. Fig.\ref{fig2} demonstrates that in challenging conditions such as cloud-covered and low-light scenes, methods~\cite{RSVQA,zhang2022domain, HRVQA} relying solely on optical images fail to extract universally meaningful features, resulting in incorrect answers. The limitations of these models underscore the necessity of developing more robust solutions that can maintain high performance across diverse imaging conditions, thereby enhancing the reliability and applicability of RSVQA systems in real-world remote sensing scenarios.

However, despite its potential, RSVQA faces significant challenges unique to remote sensing imagery. Remote sensing images captured from satellite platforms are often subject to conditions not typically encountered in natural scene imagery, presenting obstacles to effective analysis. As illustrated in Fig.\ref{fig1}, these adverse factors significantly hinder the performance of existing RSVQA models~\cite{RSVQA,rsivqa,HRVQA,prompt-RSVQA,crsvqa,bazi2022bi}. Fig.\ref{fig2} demonstrates that in challenging conditions such as cloud-covered and low-light scenes, methods~\cite{RSVQA,rsivqa, HRVQA} relying solely on optical images fail to extract universally meaningful features, resulting in incorrect answers. The limitations of these models underscore the necessity of developing more robust solutions that can maintain high performance across diverse imaging conditions, thereby enhancing the reliability and applicability of RSVQA systems in real-world remote sensing scenarios.

%Recently, there has been significant progress in the field of RSVQA, with numerous approaches being developed to tackle various challenges. Representative methods include RSVQA~\cite{RSVQA}, prompt-RSVQA~\cite{prompt-RSVQA}, CDVQA~\cite{CDVQA} and FloodNet~\cite{floodnet}. While these approaches have shown promising results on existing datasets, they primarily focus on daytime images captured under favorable conditions. The robustness of these methods in challenging scenarios, such as cloud occlusion and low illumination, remains largely unexplored. This gap underscores the need for innovative solutions that leverage complementary data sources to enhance RSVQA model performance across diverse environmental conditions.

Recently, there has been significant progress in the field of RSVQA, with numerous approaches being developed to tackle various challenges. Representative methods include RSVQA~\cite{RSVQA}, MAIN~\cite{rsivqa}, FETH~\cite{yuan2022easy}, SHRNet~\cite{zhang2023spatial}, MQVQA~\cite{crsvqa}, prompt-RSVQA~\cite{prompt-RSVQA}, SAM-VQA~\cite{SAM-VQA}, CDVQA~\cite{CDVQA} and FloodNet~\cite{floodnet}. While these approaches have shown promising results on existing datasets, they primarily focus on daytime images captured under favorable conditions. The robustness of these methods in challenging scenarios, such as cloud occlusion and low illumination, remains largely unexplored. This gap underscores the need for innovative solutions that leverage complementary data sources to enhance RSVQA model performance across diverse environmental conditions.

Synthetic Aperture Radar (SAR), with its unique all-time and all-weather imaging capabilities, offers a promising solution to mitigate the limitations of optical sensors. SAR's ability to penetrate cloud cover and capture high-quality images regardless of lighting conditions~\cite{wang2023category,liu2016deep} makes it an ideal complement to optical imagery. Optical-SAR fusion has the potential to overcome single-sensor limitations in terms of scene information content and resolution~\cite{gomez2015multimodal,schmitt2016data,mou2017multitemporal}. 

Recent advancements in optical-SAR feature-level fusion have demonstrated significant improvements in applications such as land cover classification~\cite{mcanet} and object detection~\cite{wang2023category,zhang2022domain,shi2022unsupervised}. These fusion techniques show particular promise in addressing imaging challenges posed by cloudy occlusion and low-light environments, potentially benefiting RSVQA models through improved visual-textual understanding and reasoning. Despite the promising advancements in optical-SAR fusion for various remote sensing tasks, its application in the context of RSVQA remains largely unexplored. This critical research area offers an important opportunity to develop novel, task-specific fusion approaches that could substantially advance the field of RSVQA.

In this work, we propose a novel Text-guided Coarse-to-Fine Fusion Network (TGFNet) that leverages high-level semantic connections between question text and multi-source images to guide the network towards learning complementary joint representations. Specifically, we develop a Text-guided Coarse-to-Fine Attention Refinement (CFAR) module to progressively focus on question-relevant regions while suppressing background noise in complex remote sensing scenes. Furthermore, we introduce an Adaptive Multi-Expert Fusion (AMEF) module that dynamically integrates different experts, enabling adaptive fusion of optical and SAR features. In addition, this research field still lacks a large-scale benchmark dataset with well-labeled question-answer text and well-aligned optical-SAR images, which is essential for the training and comprehensive evaluation of optical-SAR RSVQA methods. To this end, we construct the OSVQA (Optical-SAR Visual Question Answering) dataset, a new large-scale benchmark dataset comprising 6,008 aligned optical-SAR image pairs and 1,036,694 question-answer pairs across 16 diverse question types. Notably, we incorporate a unique category of questions that require the model to assess the quality and reliability of optical-SAR data for optimal source selection during reasoning. Extensive experiments on the OSVQA dataset demonstrate that our TGFNet effectively fuses complementary information from optical and SAR modalities, resulting in significant performance improvements in challenging scenarios such as low-light conditions and cloud occlusions.

In summary, the contributions of this article can be summarized as follows:\par

\begin{itemize}
\item We propose a Text-guided Coarse-to-Fine Fusion Network (TGFNet) that leverages high-level semantic relationships between questions and multi-source images to achieve a complementary fusion of key information across these images. This novel approach significantly enhances RSVQA performance under adverse imaging conditions through effective optical-SAR fusion.

\item We introduce two key modules: the Text-guided Coarse-to-Fine Attention Refinement (CFAR) module, which identifies question-relevant regions while suppressing background noise, and the Adaptive Multi-Expert Fusion (AMEF) module, which dynamically integrates features from both optical and SAR images, enabling robust feature fusion.

\item To the best of our knowledge, the proposed OSVQA dataset is the first large-scale and well-annotated optical-SAR benchmark dataset. This dataset contains 6,008 image pairs and 1,036,694 question-answer pairs, with each image pair averaging 172 complex questions across 16 different question types.

\item Extensive experimental results on our proposed dataset demonstrate that TGFNet effectively integrates the respective advantages of optical and SAR images, significantly enhancing the robustness of RSVQA under challenging imaging conditions.

\end{itemize}

\section{Related work}
This section presents an overview of recent advancements in Visual Question Answering (VQA), Remote Sensing Visual Question Answering (RSVQA), and optical-SAR image fusion, which constitute the foundation of our work.

\subsection{Visual Question Answering}
Visual Question Answering (VQA)~\cite{VQA} witnesses significant progress with the advent of deep learning techniques. Early approaches focus on the joint embedding of visual and textual features. Malinowski et al.~\cite{malinowski2015ask} propose Neural-Image-QA, utilizing recurrent neural networks for challenging image-related tasks. Gao et al.~\cite{gao2015you} introduce the mQA model, treating VQA as a classification problem by feeding feature vectors into a linear classifier. Introducing the Transformer~\cite{transformer} leads to a paradigm shift toward attention-based methods. Shih et al.~\cite{shih2016look} develop a model that learns to answer visual questions by selecting image regions relevant to text-based queries. Yu et al.~\cite{yu2019deep} propose the Deep Modular Attention Network, employing self-attention units for intra-text interactions and guided attention units for text-image interactions. Zeng et al.~\cite{zeng2021multi} introduce X-VLM, a multi-granularity vision-language pre-training approach that reconstructs existing datasets into visual concepts and corresponding texts. To address complex and diverse questions, Wang et al.~\cite{wang2015explicit} propose Ahab, a VQA method that infers image content using large-scale knowledge bases. Wu et al.~\cite{wu2016ask} develop a method that constructs textual representations of image semantic content and merges them with textual information from knowledge bases, enhancing scene understanding and enabling broader question-answering capabilities.

While these VQA methods show promising results, they are primarily designed for natural scene images and often struggle with the multi-scale features and complex semantics inherent in remote sensing imagery.

\subsection{Remote Sensing Visual Question Answering}
RSVQA~\cite{RSVQA} extends VQA techniques to the domain of remote sensing, aiming to efficiently interpret rich geospatial information and relationships in satellite and aerial imagery. However, RSVQA encounters unique challenges due to complex backgrounds, significant scale variations, and sensitivity to lighting conditions inherent in remote sensing data. To address these challenges, researchers propose various solutions. Zheng et al.~\cite{rsivqa} employ attention mechanisms to align image regions with query words and use bilinear fusion to generate joint representations of image-question pairs. Yuan et al.~\cite{yuan2022easy} integrate regional and global information to obtain multi-scale image representations and employ self-paced curriculum learning to train models from easy to difficult questions. Recent works leverage pre-trained models and multi-scale reasoning. Bazi et al.~\cite{bazi2022bi} utilize a pre-trained CLIP~\cite{clip} network for effective embedding of images and questions, capturing intricate intra-modal and inter-modal connections. Zhang et al.~\cite{zhang2023spatial} propose a method that uses textual information to guide visual-spatial reasoning across multiple scales. Zhang et al.~\cite{crsvqa} introduce a multistep question-driven approach, repeatedly focusing on the image through an attention mechanism for detailed inference. Additionally, new RSVQA tasks are being proposed for specific scenarios. Sarkar et al.~\cite{SAM-VQA} introduce a disaster assessment QA dataset, enhancing RSVQA performance in disaster evaluation. Yuan et al.~\cite{CDVQA} explore VQA applications in remote sensing change detection using multi-temporal aerial imagery. 

Despite these advancements, existing RSVQA methods primarily rely on optical images captured under favorable conditions. They often struggle with extreme imaging conditions such as low-light and cloud-covered scenes, limiting their applicability in real-world scenarios. 

\subsection{Optical and SAR Image Fusion}
Optical-SAR image fusion is a critical area in remote sensing that aims to combine the complementary strengths of both modalities, thereby enhancing information content and improving the robustness of image analysis across diverse environmental conditions. Fusion methods can be broadly categorized into traditional approaches and deep learning-based techniques. Deep learning methods, primarily utilizing Convolutional Neural Networks (CNNs)~\cite{Alexnet} and Generative Adversarial Networks (GANs)~\cite{GAN}, show promise in addressing challenges such as large imaging discrepancies and spectral mismatches between optical and SAR imagery. Li et al.~\cite{li2020multimodal} and Ienco et al.~\cite{ienco2019combining} propose multi-channel, multi-branch networks for multi-scale feature extraction. He and Yokoya~\cite{he2018multi} expand inputs to multi-temporal optical and SAR images, employing nonlinear residual networks for fusion. To bridge the modality gap, several GAN-based approaches are proposed. Gao et al.~\cite{gao2019cloud} employ GANs to transform SAR images into an optical format to fill cloudy regions in optical images. Fu et al.~\cite{fu2021reciprocal} introduce a multi-level cascaded residual connection GAN framework for mutual transformations between optical and SAR images. Grohnfeldt et al.~\cite{grohnfeldt2018conditional} develop SAR-OPT-CGAN, a conditional GAN specifically designed for SAR and multispectral image fusion.

The fusion of optical and SAR images shows the potential to enhance downstream remote sensing tasks. Li et al.~\cite{mcanet} achieve a 5\% increase in land cover classification accuracy using fused optical-SAR images compared to methods based solely on optical images. Wang et al.~\cite{wang2023category} enhance SAR target detection by transferring location knowledge from optical images through knowledge distillation. 

While optical-SAR fusion demonstrates promising results in various remote sensing applications, its potential in the context of RSVQA remains largely unexplored. Leveraging the complementary strengths of optical and SAR modalities could potentially address the limitations of current RSVQA methods, particularly under challenging imaging conditions.

\begin{figure*}
    \centering
    \includegraphics[width=.94\textwidth]{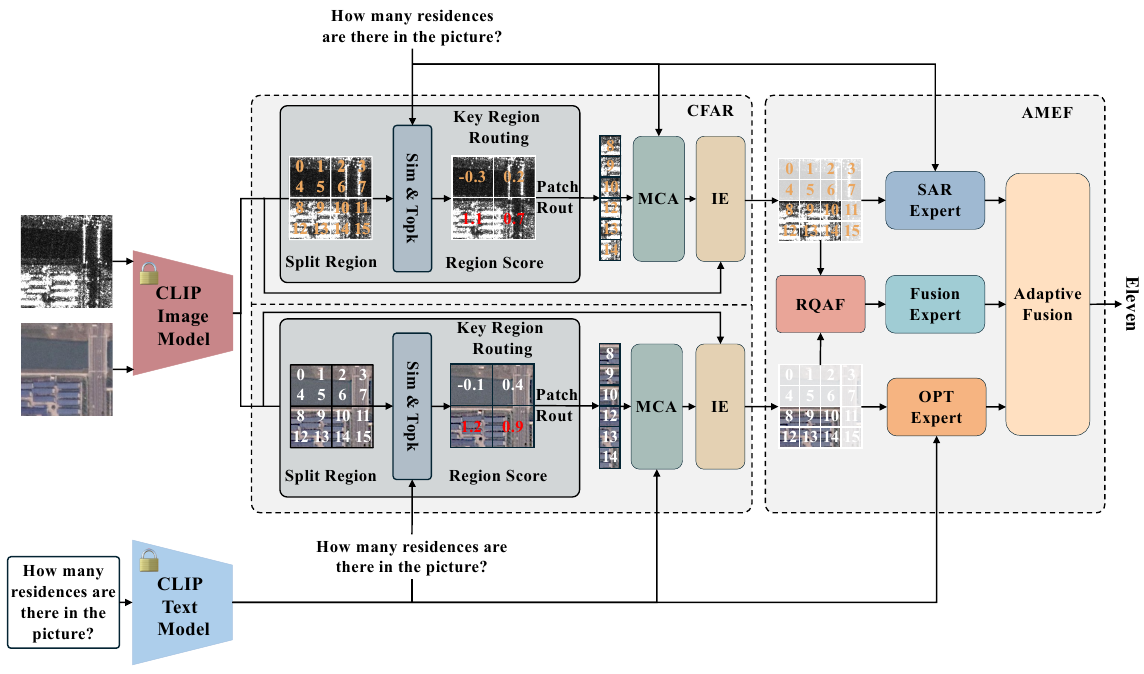}
    \caption{
    The overall framework of TGFNet is as follows: First, the CLIP~\cite{clip} model fine-tuned on OSVQA dataset is employed for initial feature extraction from both text and images. Next, we propose the Text-guided Coarse-to-Fine Attention Refinement (CFAR) module, which consists of two identical structures, each comprising KRR, MCA, and IE. This module is designed to focus on the image regions relevant to the given question. To effectively leverage the complementary strengths of SAR, optical, and fusion images for answer prediction, we introduce the Adaptive Multi-Expert Fusion (AMEF) module, which includes the SAR Expert, OPT Expert, Fusion Expert, RQAF, and AF.
    }
    \label{fig3}
\end{figure*}

\section{Methodology}
The overall framework of TGFNet is discussed in Section \ref{overallframework}. The proposed CFAR and AMEF modules are presented in Sections \ref{method-B} and \ref{method-C}, respectively. The loss function of TGFNet is detailed in Section \ref{loss}.\par

\subsection{Overall Framework}
\label{overallframework}

Visual information plays a crucial role in VQA tasks. However, optical images are prone to information loss under adverse conditions, such as low-light or cloud-covered scenes, which significantly impairs VQA performance. To address this challenge, we propose TGFNet, a novel framework that leverages the complementary strengths of optical and SAR images through an effective Coarse-to-Fine fusion strategy, thereby mitigating the impact of challenging imaging conditions on VQA tasks.
As illustrated in Fig.\ref{fig3}, our TGFNet comprises two main components:  

\textbf{(1) Text-guided Coarse-to-Fine Attention Refinement (CFAR) Module.} This module directs attention from broad areas to finer details in the optical-SAR images based on the semantic correlation between the question and the images. The identified key regions are subsequently used to enhance the optical-SAR images further.  

\textbf{(2) Adaptive Multi-Expert Fusion (AMEF) Module.} To leverage the unique strengths of different image modalities, this proposed module processes SAR images, optical images, and fused images through their respective specialized experts.

% The processing flow of TGFNet begins with a pre-trained CLIP~\cite{clip} model, which encodes text-based questions, optical images, and SAR images into a unified feature space. This unified representation facilitates effective cross-modal interactions in the subsequent stages. The CFAR module then identifies and enhances question-relevant regions in both optical and SAR images. These enhanced regions are subsequently fed into the AMEF module to integrate information from the individual modalities and their fusion experts to generate the final answer.

The processing flow of TGFNet begins with a pre-trained CLIP~\cite{clip} model, which is fine-tuned on our proposed OSVQA dataset. The model encodes text-based questions, optical images, and SAR images into a unified feature space. This unified representation facilitates effective cross-modal interactions in the subsequent stages. The CFAR module then identifies and enhances question-relevant regions in both optical and SAR images. These enhanced regions are subsequently fed into the AMEF module to integrate information from the individual modalities and their fusion experts to generate the final answer.

By integrating these components, TGFNet effectively utilizes complementary information from optical and SAR images, guided by question semantics. This progressive refinement approach enables robust performance in challenging RSVQA scenarios by focusing on increasingly relevant and detailed information throughout the processing pipeline.

\subsection{Text-guided Coarse-to-Fine Attention Refinement}
\label{method-B}

Remote sensing images contain diverse ground object information, but only a subset is relevant for answering questions. To address this challenge, we propose a Text-guided Coarse-to-Fine Attention Refinement (CFAR) module, which consists of three essential components: Key Region Routing (KRR), Multi-head Cross-Attention (MCA), and Image Feature Enhancement (IE).  

The core idea of the proposed module is to filter out irrelevant region pairs at a coarse level, retaining only a small subset of routed regions for further processing. Subsequently, finer-grained token-level attention is applied to refine the image details. This process can be summarized as follows:

\begin{figure}
    \centering
    \includegraphics[width=8cm]{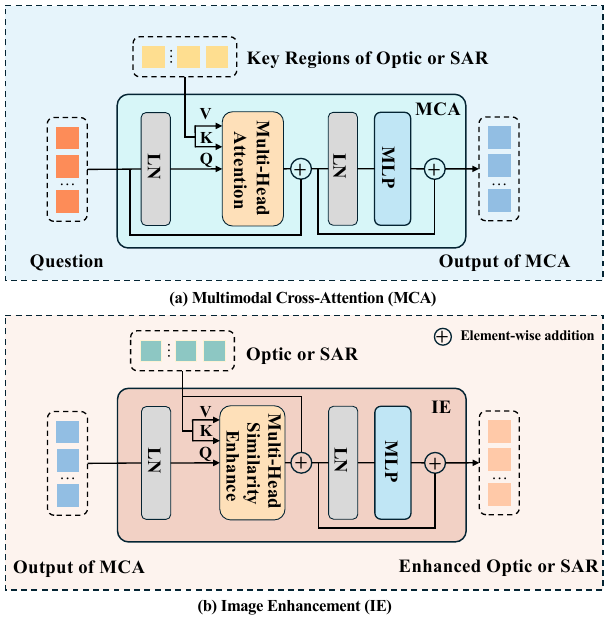}
    \caption{
    The network structures of MCA and IE are illustrated as follows. Panel (a) shows the structure of the MCA, which consists of two LayerNorm layers, a multi-head cross-attention layer, and a Multi-Layer Perceptron (MLP). It takes key regions of optical or SAR images and the question representation as inputs, outputting their fused results. Panel (b) depicts the structure of the IE, which comprises two LayerNorm layers, a Multi-Head Similarity Enhancement layer, and an MLP. The IE enhances the input optical and SAR images using the output from the MCA, highlighting regions relevant to the question.
    }
    \label{fig4}
\end{figure}

\subsubsection{Key Region Routing.}
\label{key region routing}
During the feature extraction stage, the question features, SAR image features and optical image features obtained from the CLIP Text Model and CLIP Vision Model are denoted as $F_q \in \mathbb{R}^{B \times N \times D}$, $F_{s} \in \mathbb{R}^{B \times M \times D}$ and $F_{o} \in \mathbb{R}^{B \times M \times D}$, respectively, where $B$, $N$, $M$ and $D$ represent the batch size, question length, number of image patches and feature embedding dimensionality. The optical image \( F_o \) is initially divided into \( T \) regions, each containing \( P \) patches. The average pooling of these \( P \) patch features represents the corresponding region, resulting in a coarse region representation of the optical image, denoted as \( F_{or} \in \mathbb{R}^{B \times T \times D} \). Next, by computing the correlation between the question representation \( F_q \) and each region in the optical image representation \( F_{or} \), we obtain the question-region correlation scores \( S \in \mathbb{R}^{B \times T} \) as follows:

\begin{equation}
\label{Eq1}
\begin{split}
S=Mean(\frac{(W_TF_q)(W_IF_{or})^T}{\sqrt{D}})
\end{split}
\end{equation}

where the $Mean$ refers to taking the average along dimension 1, $W_T \in \mathbb{R}^{D\times D}$ and $W_I \in \mathbb{R}^{D\times D}$ are learnable matrices.
Using the correlation scores \( S \), the top \( k \) regions with the highest scores are selected as key regions for the optical image. The codebook, which maps regions to their patches, is then used to retrieve the corresponding patch representations from \( F_{o} \), resulting in key region representations denoted as \( F_{ok} \in \mathbb{R}^{B \times K \times D} \), where \( K = P \times k \). A similar procedure is applied to the SAR image, yielding key region representations denoted as \( F_{sk} \in \mathbb{R}^{B \times K \times D} \).

\subsubsection{Multi-head Cross-Attention.}
The key region representation \( F_{ok} \) and the question representation \( F_q \) are first fed into the MCA. As shown in Fig.\ref{fig4}, the MCA consists of two LayerNorm layers, two residual connections, a multi-head cross-attention layer, and a Multi-Layer Perceptron (MLP). The MLP is a simple feedforward network with two fully connected layers and a Gaussian Error Linear Unit (GELU) activation between them. In the multi-head cross-attention layer, \( F_q \) and \( F_{ok} \) are projected using the learnable weight matrices \( W_Q \), \( W_K \) and \( W_V \), resulting in \( Q = W_Q F_q \), \( K = W_K F_{ok} \) and \( V = W_V F_{ok} \). The multi-head cross-attention layer uses eight parallel self-attention heads to compute the scaled dot-product similarity between \( Q \), \( K \) and \( V \), generating attention scores as follows:

\begin{equation}
\label{Eq2}
Attention(Q,K,V) = softmax\left(\frac{QK^T}{\sqrt{D}}\right)V
\end{equation}

The outputs of all heads are concatenated and projected through another learnable weight matrix. The resulting representation \( F_{qo} \) is computed as follows:

\begin{equation}
\label{Eq3}
\begin{split}
   F_{qo} = MLP(LN(Attention(Q,K,V)))
\end{split}
\end{equation}

Next, \( F_{qo} \) and the optical image representation \( F_o \) are input into the IE to produce the key region-enhanced optical image representation \( F_{OE} \in \mathbb{R}^{B \times M \times D} \).

\subsubsection{Image Feature Enhancement.}
\label{Image Feature Enhancement.}
The IE, similar to the MCA, replaces the multi-head cross-attention layer with a Similarity Enhancement (SE) layer. The SE layer, which also has eight independent heads,
processes \( F_{qo} \) and \( F_o \) to compute $Q$, $K$ and $V$ as \( Q = W_Q F_{qo} \), \( K = W_K F_o \) and \( V = W_V F_o \). Each head first computes the similarity between \( Q \) and \( K \), averages the similarity scores, and then performs pointwise multiplication with \( V \) as follows:

%The IE , similar to the MCA, replaces the multi-head cross-attention layer with a Similarity Enhancement (SE) layer. The SE layer, which also has eight independent heads, processes \( F_{qo} \) and \( F_o \) to compute \( Q \), \( K \), and \( V \). Each head first computes the similarity between $Q$ and $K$, averages along the first dimension and then performs pointwise multiplication with $V$ as follows:
%followed by pointwise multiplication between the averaged similarity scores and \( V \).

%The IE, similar to the MCA, includes two LayerNorm layers, two residual connections, a Similarity Enhancement (SE) layer and an MLP. The MLP is a two-layer feedforward network with GELU activation. The SE layer processes the outputs from the MCA, \( F_{qo} \), and \( F_o \). Using learnable weight matrices, it computes $Q$, $K$, and $V$ as \( Q = W_Q F_{qo} \), \( K = W_K F_o \), and \( V = W_V F_o \). The SE layer consists of eight independent heads. Each head first computes the similarity between $Q$ and $K$, averages along the first dimension, and then performs pointwise multiplication with $V$.

\begin{equation}
\label{Eq4}
\begin{split}
   SE(Q,K,V) = Softmax\left(Mean\left(\frac{QK^T}{\sqrt{D}}\right)\right) \ast V
\end{split}
\end{equation}

The outputs of all heads are concatenated and projected through an additional learnable weight matrix. Then, $F_{OE}$ is obtained as follows:
\begin{equation}
\label{Eq5}
\begin{split}
    F_{OE}=MLP(LN(SE(Q,K,V))
\end{split}
\end{equation}
By following the same procedure, but replacing the optical image representation with the SAR image representation, the enhanced SAR image $F_{SE}$ can be derived.

\subsection{Adaptive Multi-Expert Fusion}
\label{method-C}

To effectively leverage the complementary strengths of optical and SAR images, we propose an Adaptive Multi-Expert Fusion (AMEF) module that employs a novel two-stage adaptive fusion strategy: a Regional Quality-Aware Fusion (RQAF) module for constructing the Fusion Expert, followed by an adaptive modality experts fusion mechanism that leverages an Adaptive Fusion (AF) network to integrate predictions from the Optical Expert, SAR Expert and Fusion Expert to generate a robust final prediction.

\begin{figure}
    \centering
    \includegraphics[width=8cm]{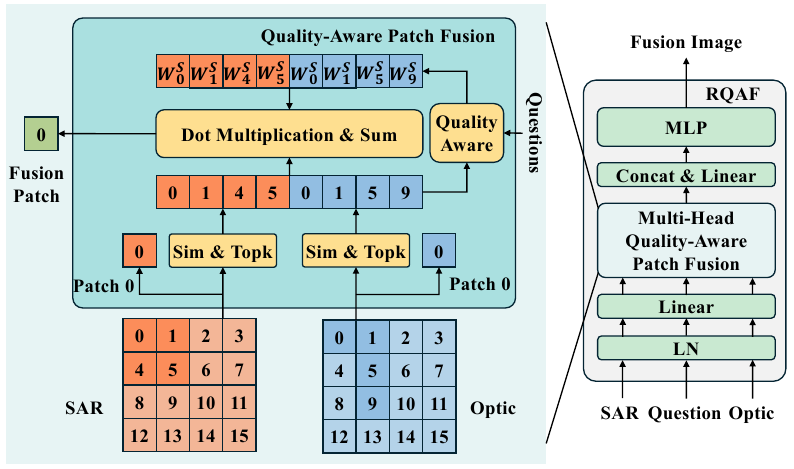}
    \caption{
    The network structure of RQAF comprises a LayerNorm layer, two linear layers, a multi-head quality-aware patch fusion layer, and an MLP. The question and optical-SAR images are simultaneously fed into the RQAF model, where the high-level semantics of the question guide the quality-aware fusion of the optical and SAR images at each spatial location.}
    \label{fig5}
\end{figure}

\subsubsection{Regional Quality-Aware Fusion.}
 Questions provide more focused semantic information than images, enabling the accurate identification of important parts in complex image scenes. This capability can also be extended to evaluate the importance of different image modalities. As illustrated in Fig.\ref{fig5}, the proposed RQAF module leverages this advantage by utilizing question semantics to guide the fusion of optical and SAR features. 

Specifically,  the enhanced SAR, optical, and question representations are mapped using three learnable weight matrices. For each spatial location \( i \in [1,2,..., M] \), the top \( R \) patches with the highest feature similarity in both optical and SAR images are selected, forming two sets: \( SET_{si} \) and \( SET_{oi} \). Utilizing the question representation \( F_q \), the feature quality scores \( SQ_{i} \) for the union of these sets are then computed. A softmax operation is applied to compute the fusion weights \( W_{i} \) for these patches. Finally, a weighted sum of the patches is performed, followed by normalization and an MLP, to generate the fused representation at location \( i \). Repeating this process across all locations yields the complete fused image representation \( F_{OS} \), which serves as the visual input for the Fusion Expert in the next stage. The calculation of \( W_{i} \) is as follows:

\begin{equation}
\label{Eq6}
\begin{split}
W_{i}=Softmax(Mean(\frac{F_qSet_i^T}{\sqrt{D}}))
\end{split}
\end{equation}
where \( Set_i \) is the union of \( SET_{si} \) and \( SET_{oi} \).

\begin{figure}
    \centering
    \includegraphics[width=9cm]{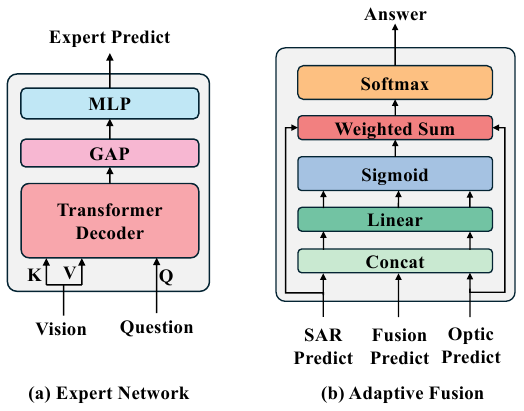}
    \caption{
     The network structures of the SAR Expert, Optical Expert, Fusion Expert, and Adaptive Fusion are described as follows. Panel (a) illustrates the structure of the three experts, each of which shares an identical architecture. Each expert receives a specific image modality representation and a text representation as inputs to predict an answer. Panel (b) illustrates the AF, which adaptively generates fusion weights based on the predictions of the three experts. These weights are subsequently used to integrate the predictions, yielding the final answer.
    }
    \label{fig6}
\end{figure}

\subsubsection{Adaptive Fusion of  Optical Expert, SAR Expert and Fusion
Expert.} 
Each expert consists of a vision-language fusion reasoning module and a classification network, as shown in Fig.\ref{fig6}. Given the importance of fine-grained interactions in vision-language fusion reasoning, we employ a two-layer Transformer Decoder network as the vision-language fusion reasoning module. The classification network is an MLP comprising two fully connected layers, with a ReLU activation function applied between them. The prediction process for each modality expert is mathematically expressed as follows:

\begin{equation}
\label{Eq9}
\begin{gathered}
P = FC(ReLU(FC(GAP(Decoder(Q, K, V)))))
\end{gathered}
\end{equation}

where $Q$, $K$, and $V$ represent the Query, Key, and Value, respectively. The Query($Q$) is derived from the question representation using a learnable weight matrix, while the Key($K$) and Value($V$) are derived from the visual representations $F_{o}$, $F_{s}$ or $F_{OS}$. $GAP$ stands for the global average pooling operation. $Decoder$ represents the Transformer Decoder network. The predictions of the Optical Expert, SAR Expert and Fusion Expert are denoted as \( P_{OE}\in \mathbb{R}^{B \times 1 \times C} \), \( P_{SE}\in \mathbb{R}^{B \times 1 \times C} \) and \( P_{OS} \in \mathbb{R}^{B \times 1 \times C}\), respectively, where \( C \) is the number of answer classes.

%Next, the predictions from the Optical Expert, SAR Expert, and Fusion Expert  to generate the final answer. 
The predictions from the Optical Expert, SAR Expert, and Fusion Expert are subsequently integrated by the AF network to produce the final answer. Initially, the predictions from the three experts are concatenated to form the input for the AF network. This input is then processed through a fully connected layer, followed by a sigmoid function, to determine the fusion weights for each expert. A weighted sum of all expert predictions is then calculated using these fusion weights. Finally, this sum is passed through a softmax function to generate the final predicted answer. The mathematical expression for this process is as follows:

\begin{equation}
\label{Eq10}
\begin{gathered}
W_{A} = Sigmoid\left(FC\left(Concat_{2}\left(P_{OE}, P_{SE}, P_{OS}\right)\right)\right) \\
Pre = Softmax\left(W_{A}\left(Concat_{1}\left(P_{OE}, P_{SE}, P_{OS}\right)\right)\right)
\end{gathered}
\end{equation}
where $W_A\in \mathbb{R}^{B \times 1 \times 3}$ denotes the adaptive fusion weights for the Optical Expert, SAR Expert, and Fusion Expert. $Pre$ represents the final prediction result of the model and \( Concat_i \) denotes concatenation along the $i$ dimension.

This multi-expert network design, combined with our adaptive fusion mechanism, enables our model to effectively leverage complementary information from different image types while dynamically adapting to diverse question types and challenging image conditions. The specialized experts allow for in-depth analysis of each image type, while the adaptive fusion ensures optimal integration of their insights, leading to more robust and accurate RSVQA performance.

\setlength{\floatsep}{5pt plus 2pt minus 2pt}
\setlength{\textfloatsep}{5pt plus 2pt minus 2pt}
\setlength{\intextsep}{5pt plus 2pt minus 2pt}
\subsection{Loss Function}
\label{loss}

We employ standard cross-entropy loss as the overall network constraint, with batch-based training involving all three experts and the AMEF module. During inference, however, only the AMEF module’s output, which is a dynamically weighted sum of the three experts’ predictions, is used for answer prediction. The loss function for each expert, along with their adaptive integration, is defined as follows:

\begin{equation}
\label{Eq11}
\begin{split}
\hat{\mathcal{L}}= -\frac{1}{N_{train}}\sum_{i=1}^{N_{train}}y_ilog\hat{y_i}
\end{split}
\end{equation}
The total loss over the training set is given as follows:

\begin{equation}
\label{Eq12}
\begin{split}
\mathcal{L}= \lambda_1\hat{\mathcal{L}}_{OE}+\lambda_2\hat{\mathcal{L}}_{SE}+\lambda_3\hat{\mathcal{L}}_{OS}++\lambda_4\hat{\mathcal{L}}_{TGF}
\end{split}
\end{equation}
where $\hat{\mathcal{L}}_{OE}$, $\hat{\mathcal{L}}_{SE}$, $\hat{\mathcal{L}}_{OS}$ and $\hat{\mathcal{L}}_{TGF}$ represent the prediction losses of the Optical Expert, SAR Expert, Fusion Expert and their integrated predictions, respectively. $\lambda_1$, $\lambda_2$, $\lambda_3$ and $\lambda_4$ are regularization parameters that control the contribution of each loss. In this study, all parameters are set to $0.5$.

\begin{figure*}
    \centering
    \includegraphics[width=\textwidth]{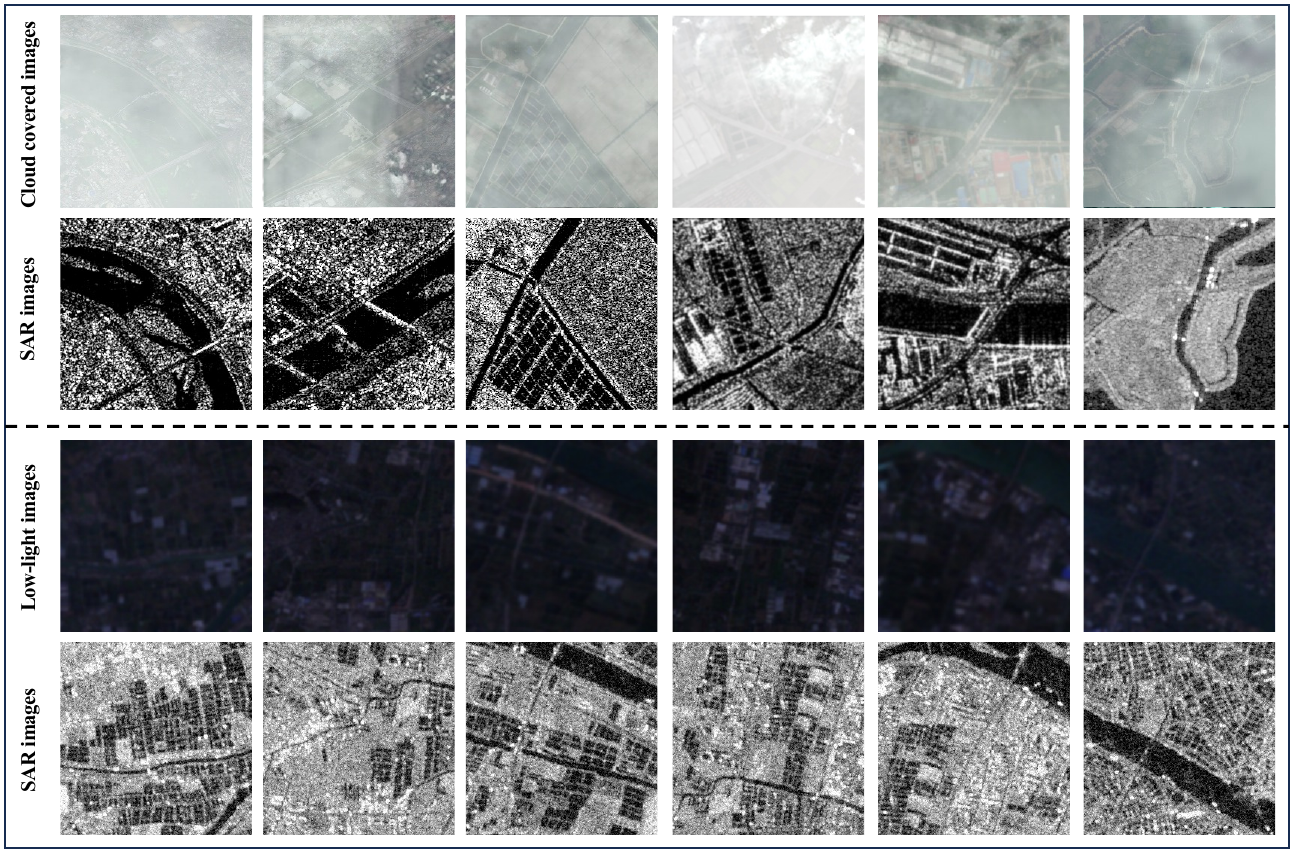}
    \caption{
    Optical and SAR image pairs in OSVQA. The first two rows display optical and SAR pairs under cloud-covered conditions, while the last two rows show optical and SAR pairs under low-light conditions. The scenes depicted in the optical and SAR images within the same column are identical.
    }
    \label{fig7}
\end{figure*}

\section{DATASET}
This section presents an analysis of existing RSVQA datasets, followed by a detailed description of the construction process, statistical analysis, and the challenges associated with the proposed OSVQA dataset.
\subsection{Existing Datasets}

To advance the development of RSVQA, researchers introduce a range of influential datasets, categorized into general-purpose and special-purpose question-answer datasets. The general-purpose datasets aim to enhance the understanding of generic geospatial information and their relationships within remote sensing images, while the special-purpose datasets are designed to promote RSVQA research in specific remote sensing scenarios, such as change detection and disaster assessment. Below, we detail several of the most representative RSVQA datasets.
%In the following, we detail several of the most representative RSVQA datasets.

\noindent\textbf{RSVQA}~\cite{RSVQA} addresses the scarcity of RSVQA datasets by introducing RSVQA-LR and RSVQA-HR. Generated using an automated template-based approach, RSVQA-HR contains 1,066,316 Q\&A pairs, while RSVQA-LR includes 77,232 pairs, marking a significant advancement in this field.

\noindent\textbf{RSIVQA}~\cite{rsivqa} comprises 37,264 images and 111,134 Q\&A pairs, featuring 91 distinct questions and 574 unique answers. This dataset is constructed using a combination of manual annotation and automated generation, contributing significantly to scene understanding and object recognition in remote sensing.

\noindent\textbf{CRSVQA}~\cite{crsvqa} is a manually annotated dataset containing 4,639 images across 30 scenes, with 4,644 questions (674 unique) and 327 distinct answers. Its complex question formats are designed to enhance model inference capabilities and improve understanding of intricate geospatial information.

\noindent\textbf{FloodNet-VQA}~\cite{floodnet} focuses on VQA tasks related to buildings, roads, and overall image assessment in disaster scenarios. It provides over 4,500 question-image pairs, averaging 3.5 questions per image, categorized into four types: simple counting, complex counting, overall condition recognition, and yes/no questions.

\noindent\textbf{CDVQA}~\cite{CDVQA}, designed for change detection in RSVQA, includes 2,968 pairs of pre- and post-change aerial images. It contains over 122,000 automatically generated Q\&A pairs covering five question types: change detection, increase/decrease, extent of change, maximum/minimum change, and rate of change.

While these datasets have advanced RSVQA research in various domains, they primarily focus on well-imaged optical remote sensing data, neglecting the challenges posed by adverse lighting and weather conditions. To address this limitation, we propose OSVQA, an RSVQA dataset combining optical and SAR images to enhance robustness in challenging scenarios, as shown in Fig.\ref{fig7}.

\begin{table}[!ht]
\renewcommand{\arraystretch}{1.3}
\begin{center}
\caption{Image attribute categories and levels used for dataset annotation.}
\resizebox{0.5\textwidth}{!}{
\begin{tabular}{c|c|c}
\hline
\textbf{Image Attributes}& \textbf{Levels}& \textbf{Attributes} \\ \hline

\multirow{6}{*}{Intrinsic Attributes}& \multirow{2}{*}{Image}& Match, Mist-Dark or Not, Urban or Not\\
 & &Residential or Not\\ \cline{2-3} 
                                      & \multirow{4}{*}{Object}& Land Cover, Subcategory, Number\\  
                                      &                                & Location, Shape, Area\\ 
                                      &                                & Length, Distribution\\
 & &Quality\\ \hline

\multicolumn{2}{c|}{Relational Attributes} & Relative Positions, Relative Distances \\ \hline

\end{tabular}
}
\label{table1}
\end{center}
\end{table}

\subsection{Dataset Construction}
\textbf{Image Collection.}
We construct the OSVQA based on the existing QXS-SAROPT~\cite{QXS-SAROPT} and OGSOD-1.0~\cite{wang2023category}. The QXS-SAROPT, designed to advance deep learning in remote sensing, contains 20,000 pairs of high-resolution (1-meter) optical-SAR images, each 256$\times$256 pixels. OGSOD-1.0, aimed at multi-target detection using optical-guided SAR images, includes 18,331 pairs with a resolution of 10 meters, also 256$\times$256 pixels each. From the aforementioned datasets, we selected 3,008 and 3,000 pairs of approximately aligned optical-SAR images, respectively, for further annotation.\par
%From these datasets, We selected 3,008 and 3,000 pairs of approximately aligned optical-SAR images for further annotation.

\textbf{Image Annotation.}
To combine the efficiency of template-generated annotations with the accuracy of manual annotations, we employ a two-stage semi-automatic method to generate questions and answers. Initially, we use a custom annotation tool for manually annotating each image, capturing attributes such as land cover types, locations, relative positions, quantity, distributions, shapes, and overall scene descriptions to ensure a comprehensive summary of the image content. Specific attribute categories are detailed in Table\ref{table1}. Based on this attribute information, we then use predefined templates to automatically generate a diverse set of complex reasoning questions, including those about relative positions, quantities, lengths, and areas. The following describes the processes for generating questions and answers, as well as the methods for image processing and dataset partitioning.\par

\textbf{Question Generation.} After completing the image attribute annotation, we automatically generate questions based on these attributes. For image-level intrinsic attributes, we establish four question categories: "match", "fog-dark", "urban" and "theme". These questions address image alignment, the presence of clouds or darkness, and whether the scene is urban or residential, with answers derived directly from the annotations.

\begin{figure*}
    \centering
    \includegraphics[width=\textwidth]{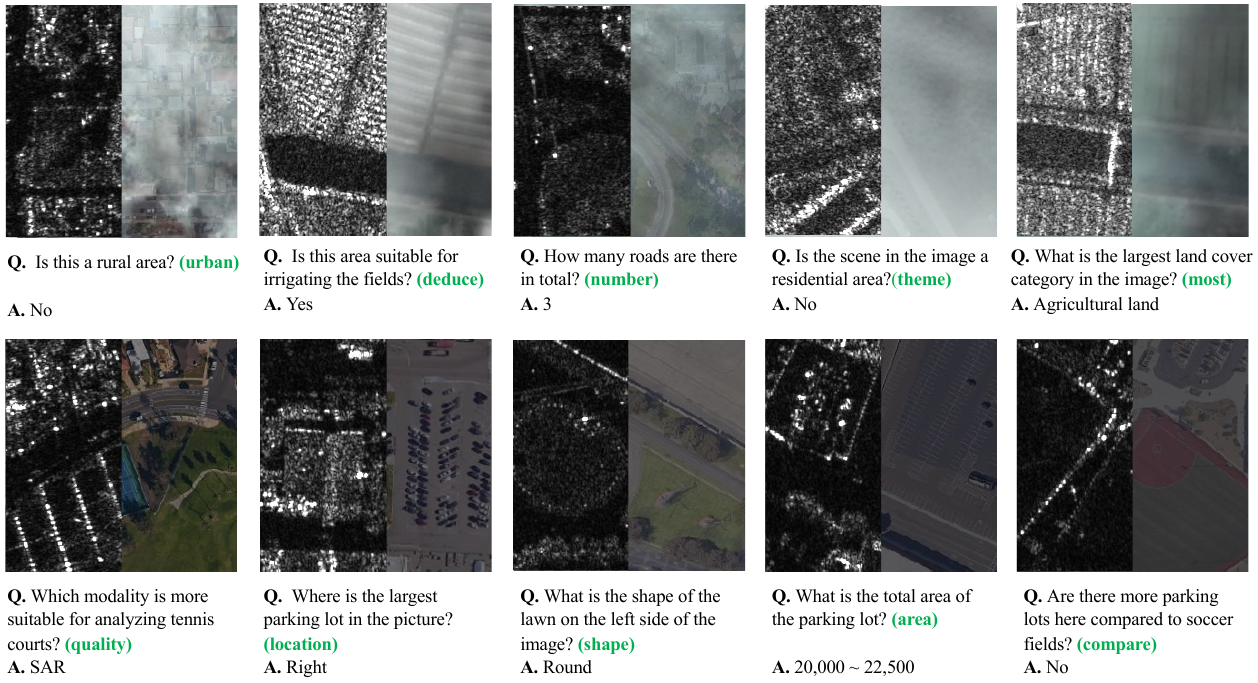}
    \caption{
    Visualization examples from the generated OSVQA dataset. We present ten data samples, each consisting of a SAR image on the left and the corresponding optical image under cloud-covered or low-light conditions on the right. A question and its corresponding answer accompany each sample.
    }
    \label{fig8}
\end{figure*}

For object-level attributes, we pose eight question categories regarding land cover and its associated subcategories: "number", "presence", "location", "shape", "area", "length", "distribution" and "quality" (unique optical-SAR modality quality assessment category). Additionally, we generate comparative questions based on "number", "area" and "length". To enhance model inference, we include manually annotated reasoning questions ("deduce") to capture subtle, complex inferences. Various question types in OSVQA are presented in Fig.\ref{fig8}. Ultimately, based on 6,008 optical-SAR image pairs, we have created the OSVQA dataset with 16 question types and 1,036,694 question-answer pairs.\par

\textbf{Image Processing.} The OSVQA dataset comprises both real-world and simulated challenging conditions. To simulate adverse lighting and weather conditions, we selectively adjust certain images through appropriate artificial modifications. For the simulated portion, we annotate the "fog-dark" and "quality" attributes post-processing. This combination of real and simulated data enhances the dataset's utility for developing robust RSVQA models capable of performing well under various environmental conditions.\par

\textbf{Dataset Segmentation.} For our dataset, we split the 6,008 optical-SAR image pairs into training, testing, and validation sets in a 3:1:1 ratio, resulting in 3,602, 1,204, and 1,202 images, respectively. Correspondingly, the question-answer pairs are distributed as 625,086 for training, 208,578 for testing, and 203,030 for validation. The proportions of rural and urban scenes, as well as the distribution of challenges within these two scene types, are consistent across the training, testing, and validation sets.\par

\begin{figure}
    \centering
    \includegraphics[width=8.8cm]{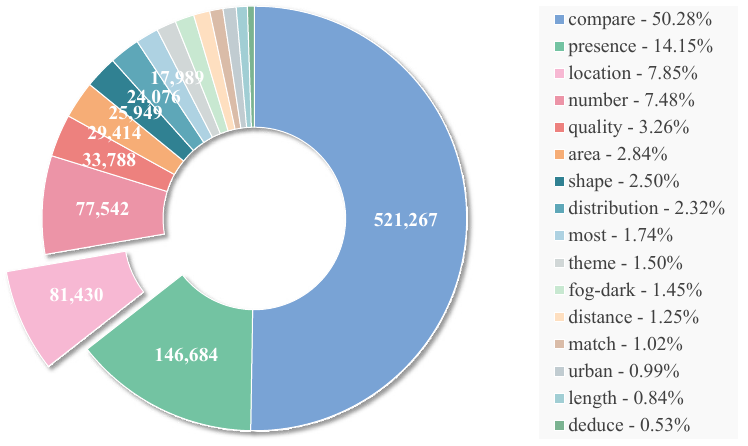}
    \caption{Visualization of the distribution of all 16 question types. Among them, the "compare" and "presence" question types have the highest proportions, while the "length" and "deduce" question types have the lowest proportions.}
    \label{fig9}
\end{figure}

\begin{table*}
\renewcommand{\arraystretch}{1.5}
\begin{center}
\begin{threeparttable}
\caption{Comparison of existing RSVQA datasets}
\setlength{\tabcolsep}{5pt}	
\scriptsize 
\begin{tabular}{cccccccccccccccccc}
\hline
\multirow{2}{*}{Dataset} &  Number of & Number & Question & Number of  & Number of & Unique Ques- & Questions per & \multirow{2}{*} {Image Type} \\
 & Images & of Scenes &  Types & Questions & Answers & tions &  Image &  \\
\hline
RSVQA-HR~\cite{RSVQA} & \uline{10,659} & \textgreater 7 & 4 & \textbf{1,066,316} & 98 & - & \uline{100.04} & Optical \\
\hline
RSVQA-LR~\cite{RSVQA} & 772 & \textgreater 7 & 4 & 77,232 & 9 & - & \uline{100.04} & Optical \\
\hline
RSIVQA~\cite{rsivqa} & \textbf{37,264} & \textbf{38} & 9 & 111,134 & \textbf{574} & 91 & 2.98 & Optical \\
\hline
CRSVQA~\cite{crsvqa} & 4,639 & 30 & 3 & 4,644 & \uline{327} & \uline{674} & 1.00 & Optical \\
\hline
FloodNet-VQA~\cite{floodnet} & 2,188 & 8 & 4 & 7,355 & 41 & 15 & 3.36 & Optical \\
\hline
FloodNet-VQA V2.0~\cite{floodnet} & 2,348 & 9 & 7 & 10,480 & 49 & 43 & 8.47 & Optical \\
\hline
\textbf{OSVQA(Ours)} & 6,008 & \uline{32} & \textbf{16} & \uline{1,036,694} & 140 & \textbf{72,195} & \textbf{172.55} & \textbf{Optical\&SAR} \\
\hline

\end{tabular}
\label{table2}
\begin{tablenotes}
        \footnotesize
        \item[]
        \begin{center}
            $\textbf{Best}$, $\uline{Second\enspace best}$. Higher metrics are better
        \end{center}
      \end{tablenotes}
    \end{threeparttable}
\end{center}
\end{table*}

\subsection{Dataset Analysis}
In this part, we present an analysis of the OSVQA dataset to demonstrate its superiority.\par
% OSVQA 数据集包含 16 类问题：比较、存在、位置、数量、面积、形状、分布、最多、主题、雾-暗、距离、匹配、城市、长度、推断，以及一类特殊设计的光-SAR模态质量评估问题。每类问题的数量从 5 456 到 521 267 不等。如图十所示，不同问题类型的分布显示，比较、存在、位置和数字类问题最多，分别占50.3%、14.1%、7.9%和7.5%，而推理类问题所占比例最小，约为0.5%。这种分布反映了所构建数据集中问题的丰富性和多样性。
\textbf{Distribution of Question Types.} %The OSVQA dataset contains 16 question types: compare, presence, location, number, area, shape, distribution, most, theme, fog-dark, distance, match, urban, length, deduce and unique optical-SAR modality quality assessment category. 
The OSVQA dataset contains 16 question types: "compare", "presence", "location", "number", "area", "shape", "distribution", "most", "theme", "fog-dark", "distance", "match", "urban", "length", "deduce" and "quality". The number of questions per type ranges from 5,456 to 521,267. As shown in Fig.\ref{fig9}, "compare", "presence", "location" and "number" questions are the most prevalent, constituting 50.3\%, 14.2\%, 7.9\% and 7.5\% of the dataset, respectively, while the "deduce" questions are the least common at 0.5\%. This distribution highlights the richness and diversity of the dataset.\par

\begin{figure}[th!]
    \centering
    \includegraphics[width=9.0cm]{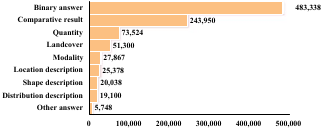}
    \caption{Visualization of the distribution of the top 30 most frequent answer categories.}
    \label{fig10}
\end{figure}

\textbf{Distribution of Answer Categories.} The OSVQA dataset includes a total of 140 distinct answer categories. Fig.\ref{fig10} shows the distribution of the 30 most frequent answer categories, which are grouped into nine types: Binary Answer, Comparative Result, Quantity, Land Cover, Modality, Location Description, Shape Description, Distribution Description, and Other Answers. Notably, semantically opposite answers, such as `yes/no', `smaller/larger', and `less/more', generally have equivalent counts. This balance highlights the even distribution of answers within the dataset.\par

\textbf{Distribution of Question Lengths.} The OSVQA dataset contains a significant number of lengthy questions. The average question length is 16.1 words, with a maximum of 69 words and a minimum of 4 words. Notably, 84.6\% of the questions exceed 20 words in length, and 12.0\% exceed 40 words. This underscores the complexity of the questions in this dataset.
%The OSVQA dataset contains a significant number of lengthy questions. The average question length is 16.12 words, with a maximum of 69 words and a minimum of 4 words. Notably, 84.56\% of the questions exceed 20 words in length and 12.01\% exceed 40 words. This underscores the complexity of the questions in this dataset.\par

\renewcommand{\arraystretch}{1.5} % 设置默认行高

% \begin{figure}[th!]
%     \centering
%     \includegraphics[width=9.0cm]{dataset/ans_length.pdf}
%     \caption{Visualization of the distribution of question lengths. The length of a question is defined by the number of words in its text.}
%     \label{fig11}
% \end{figure}

\textbf{Comparisons with Other RSVQA Datasets.} 
% 表 X 列出了与其他 RSVQA 数据集的比较，包括 RSVQA-HR、RSVQA-LR、RSIVQA、CRSVQA、FloodNet-VQA 和 FloodNet-VQA V2.0。与这些数据集相比，所提出的 OSVQA 在所有比较维度上都更胜一筹，如图像数、场景数、问题类型、问题数、答案数、唯一问题数和每幅图像注释的平均问题数。此外，OSVQA 数据集是唯一一个由光学-合成孔径雷达图像构建的 RSVQA 任务数据集，包含大量光-SAR模态质量评估问题，这意味着它可用于涉及光学-合成孔径雷达和单源光学或合成孔径雷达图像的 VQA 任务。这些比较突出表明，所构建的数据集具有场景多样性、问题类型复杂而丰富、数据规模大等特点。
Table\ref{table2} compares the proposed OSVQA dataset with the existing major RSVQA datasets. OSVQA demonstrates an advantage over these datasets in most comparison metrics: the number of scenes, question types, question numbers, unique questions, and average questions per image. Moreover, OSVQA is the only dataset constructed from optical-SAR images, featuring extensive quality assessment questions for both modalities, making it suitable for VQA tasks involving optical, SAR, or combined images. This highlights its diversity, complexity, and large scale.\par

% OSVQA 数据集共包含 1,036,694 个问题-答案对，涉及 16 种类型的问题，其中包含一类特殊设计的光-SAR模态质量评估问题。其中包括 72 195 个独特的问题、1 494 个独特的问题单词和 140 个独特的答案。共注释了 6,008 对光学合成孔径雷达图像，每对图像平均包含 172.5 个问题。通过使用半自动注释方法，OSVQA 有效和充分地挖掘了图像中的地理空间信息及其关系，为 RSVQA 任务提供了丰富多样的高质量问答数据集。

Overall, the OSVQA dataset comprises a total of 1,036,694 question-answer pairs, involving 16 types of questions. It includes 72,195 unique questions, 1,494 unique question words, and 140 unique answers. A total of 6,008 optical-SAR image pairs are annotated, each with an average of 172.55 questions. By utilizing a semi-automatic annotation approach, OSVQA effectively mines the geospatial information and its relationships within the images, providing a rich and diverse high-quality question-answer dataset for RSVQA tasks.\par

\subsection{Dataset Challenges}
Compared to existing RSVQA datasets, OSVQA introduces significant advancements, offering unique challenges and valuable contributions to the field:

\begin{itemize}
    \item \textbf{Multi-Modal Fusion for RSVQA:} By integrating optical and SAR imagery, OSVQA pioneers the exploration of multi-modal data fusion in RSVQA. This approach enables the development of algorithms capable of leveraging complementary information across modalities, particularly under challenging conditions.
    
    \item \textbf{RSVQA under Adverse Conditions:} OSVQA incorporates both real and simulated challenging conditions, including cloud cover and low-light scenarios. This feature significantly enhances the dataset's ecological validity, enabling researchers to evaluate and improve RSVQA model performance under a wide range of environmental conditions typically encountered in real-world remote sensing applications.
    
    \item \textbf{Comprehensive Question Design:} With 16 diverse question types, including novel categories for reasoning and modality quality assessment, OSVQA offers a more comprehensive evaluation framework. This expanded typology enables the assessment of models across a broader spectrum of cognitive tasks, ranging from simple recognition to complex inferential reasoning.
    
    \item \textbf{Scale and Diversity:} Comprising 6,008 optical-SAR image pairs and 1,036,694 question-answer pairs, OSVQA substantially surpasses most existing datasets in scale. The inclusion of 72,195 unique questions and 140 unique answers ensures a rich diversity of scenarios and challenges, promoting the development of more generalizable RSVQA models.

    % \item \textbf{Quality-Aware Annotations:} OSVQA uniquely incorporates quality assessment questions, enabling the evaluation of models' abilities to discern and utilize the most reliable information from different modalities. This feature is crucial for developing intelligent systems capable of making adaptive decisions under imaging conditions.
\end{itemize}

These characteristics establish OSVQA as a benchmark dataset for advancing RSVQA research, particularly in addressing the complexities of multi-modal fusion, environmental variability, and complex inferential reasoning in remote sensing applications.

\section{EXPERIMENTS}

In this section, we conduct experiments to verify the effectiveness of TGFNet on the proposed OSVQA dataset. First, we present the implementation details of the experiments in Section \ref{Implementation} and then benchmark TGFNet against state-of-the-art (SOTA) VQA models in Section \ref{Quantitative}. To further analyze the effectiveness of the core model designs, we conduct extended ablation studies in Section \ref{Ablations}.

\subsection{Implementation Details}
\label{Implementation}
All experiments are conducted in the same PyTorch environment, utilizing an NVIDIA RTX 4090 GPU. The models are trained using the Adam optimizer with parameters $\beta_{1} = 0.9$ and $\beta_{2} = 0.999$ for 100 epochs. The learning rate and batch size are set to 0.00001 and 100, respectively.

% For feature extraction, we employ a pre-trained CLIP~\cite{clip} model (clip\_vit\_base\_patch\_16) to independently extract image and text features. Specifically, the input images are scaled to a size of 224$\times$224 pixels and the SAR images are replicated along the channel dimension to form three-channel images. Additionally, the input questions are embedded into 71$\times$512 vectors to standardize the word length across all questions.

For feature extraction, an adapter module was added before the projection layer of the CLIP~\cite{clip} (clip\_vit\_base\_patch\_16) model, with other weights frozen, and fine-tuned on the OSVQA dataset using contrastive learning loss to enhance its suitability for remote sensing tasks. Building on this fine-tuned model, we utilize it for feature extraction by independently obtaining image and text features. Specifically, the input images are scaled to a size of 224$\times$224 pixels and the SAR images are replicated along the channel dimension to form three-channel images. Additionally, the input questions are embedded into 71$\times$512 vectors to standardize the word length across all questions.

For evaluation, we use the mainstream metrics of overall accuracy (OA) and average accuracy (AA). OA is defined as the ratio of correctly predicted answers to all predicted answers, while AA represents the average accuracy across all categories. To ensure a fair comparison between models with single-modality image input and those with optical and SAR multi-modal image input, we exclude the question categories "quality", "fog-dark", and "match", which involve both optical and SAR images.

Furthermore, to compare the performance of different methods using optical-SAR multi-source image input, we modify VQA methods originally based solely on optical images. For these methods, we incorporate an identical image encoder to extract features from SAR images. The optical and SAR image features are then fused using an addition operation, keeping the structure of the downstream network unchanged.

% \begin{figure*}
%     \centering
%     \includegraphics[width=\textwidth]{figure_VQA/heatmap_1pdf.pdf}
%     \caption{
%     Visualization examples of methods.
%     % 
%     }
%     \label{fig9}
% \end{figure*}

\subsection{Quantitative and Qualitative Evaluation}
\label{Quantitative}

To validate the effectiveness of our TGFNet, we build a benchmark by conducting experiments on the proposed OSVQA dataset using 6 different VQA methods, which are detailed as follows.

\textbf{RSVQA.} RSVQA~\cite{RSVQA} is a pioneering method designed for remote sensing scenarios. It combines CNNs for visual feature extraction with RNNs for natural language processing.

\textbf{MAIN.} MAIN~\cite{rsivqa} leverages convolutional features for spatial information and word vectors for semantic representation. It incorporates a mutual attention component and a bilinear model for feature fusion.

\textbf{FETH.} FETH~\cite{yuan2022easy} applies a multi-level visual feature learning approach to jointly extract language-guided holistic and regional image features. It uses a self-paced curriculum learning (SPCL) strategy with soft weighting to progressively train the model from easy to hard, based on question difficulty.

\textbf{Bi-Modal.} Bi-Modal~\cite{bazi2022bi} uses the CLIP network to embed image patches and question words. It employs attention mechanisms via dual decoders to capture intra- and interdependencies between visual and textual representations.

\textbf{HRVQA.} HRVQA~\cite{HRVQA} enhances the joint feature representation of images and questions in high-resolution aerial image VQA through a gated attention mechanism and a mutual fusion module.

\textbf{TRAR.} TRAR~\cite{trar} employs a dynamic routing scheme and a path controller module to optimize attention span selection in visual transformer layers, enhancing both global and local dependency modeling.

\textbf{RSAdapter.} RSAdapter~\cite{RSAdapter} introduces a parallel adapter and an additional linear transformation layer inserted after each FC layer within the adapter to enhance adaptability to pre-trained multimodal models.

\begin{table*}[t]
\footnotesize
\centering
\begin{threeparttable}
%\caption{Comparison of the results of our proposed method with existing RSVQA methods on OSVQA dataset. SAR refers to using SAR images as the sole visual input, OPT refers to using optical images only, and MUL refers to using both SAR and optical images as visual inputs.}
\caption{Comparison of results between our proposed method and existing RSVQA methods on the OSVQA dataset. SAR denotes using only SAR images as visual input, OPT denotes using only optical images, and MUL denotes using both SAR and optical images as inputs.}
%\label{table1}
\begin{tabular}{ccccccccccccc}
\hline
 Model & Venue & Modality & OA & AA & Compare & Precence & Locat. & Num. & Deduce & Length & Shape & Theme  \\
\hline

%SAR
RSVQA~\cite{RSVQA}    & TRGS'2020   & SAR  & 64.13 & 57.41 & 65.26& 76.67& 38.37& 68.67 & 62.98 & 47.69 & 55.29 & 81.30  \\
MAIN~\cite{rsivqa}     & TRGS'2021   & SAR  & 67.94 & 60.93 & 67.59& 80.06& 58.82& 67.35 & 73.96 & 43.87 & 54.11 & 80.51 \\
Bi-Modal~\cite{bazi2022bi} & TRGS'2022   & SAR  & 69.00 & 62.99 & 70.15& 81.89& 59.51& 69.21 & 78.59 & 45.66 & 58.95 & 83.13  \\
FETH~\cite{yuan2022easy}     & TGRS'2022   & SAR  & 65.52 & 57.65 & 65.96& 76.52& 40.68& 68.86 & 63.50 & \uline{48.25}& 55.88 & 80.64  \\
TRAR~\cite{trar}     & ICCV'2021   & SAR  & 69.11&62.89& 70.52&81.56&\uline{59.77}&68.44&78.71&45.66&57.27&83.40  \\
RSAdapter~\cite{RSAdapter}   & TRGS'2024   & SAR & 65.09 & 58.70 & 67.04 & 78.76 & 42.04 & 68.48 & 59.38 & 47.41 & 56.97 & 83.84 \\
HRVQA~\cite{HRVQA}    & ISPRS'2024  & SAR  & 68.54 & 62.26 & 69.84& 81.28& 59.75& 68.93 & 77.79 & 46.25 & 54.54 & 83.06   \\

\hline
%OPT
RSVQA~\cite{RSVQA}    & TRGS'2020   & OPT  & 64.53 & 57.81 & 65.81& 76.97& 38.77 & 68.58 & 64.00 & 47.64 & 56.88 & 82.62  \\
MAIN~\cite{rsivqa}     & TRGS'2021   & OPT  & 67.19 & 60.08 & 67.56& 79.74& 58.04 & 67.79 & 74.68 & 43.30 & 55.01 & 82.90  \\
Bi-Modal~\cite{bazi2022bi} & TRGS'2022   & OPT  & 68.98 & 62.79 & 70.80& 82.41& 56.89 & 68.89 & 78.05 & 45.45 & 58.78 & \uline{87.30}\\
FETH~\cite{yuan2022easy}      & TGRS'2022   & OPT  & 66.13 & 57.84& 66.57 & 76.60& 40.40 & 69.17& 62.87 & 47.95& 55.85& 81.55  \\
TRAR~\cite{trar}      & ICCV'2021   & OPT  & 69.87&63.33& \uline{71.30}& 83.26& 59.62 & 69.51& 79.01 & 45.72& 57.85 & 85.88  \\
RSAdapter~\cite{RSAdapter}   & TRGS'2024   & OPT & 65.19 & 58.82 & 66.69 & 77.95 & 41.21 & 69.02 & 62.85 & 46.76 & 56.74 & 84.89 \\
HRVQA~\cite{HRVQA}    & ISPRS'2024  & OPT  & 68.60 & 61.88& 70.08& 81.61 & 59.43 & 68.78 & 77.09 & 44.39 & 54.76 & 85.91  \\

\hline
RSVQA~\cite{RSVQA}    & TRGS'2020   & MUL  & 60.50 & 53.43 & 61.73& 71.60& 35.47 & 67.82 & 60.54 & \textbf{48.64} & 53.04 & 73.54  \\
MAIN~\cite{rsivqa}     & TRGS'2021   & MUL  & 68.51 & 61.66 & 68.39& 81.08& 57.48& 67.87 & 75.49 & 45.48 & 56.29 & 83.54  \\
Bi-Modal~\cite{bazi2022bi} & TRGS'2022   & MUL  & \uline{69.97} & \uline{64.15} & 71.06& \uline{83.84}& 59.02& \uline{69.68}& \textbf{80.71} & 45.57 & \uline{59.98}& 86.07  \\
FETH~\cite{yuan2022easy}      & TGRS'2022   & MUL  & 60.46 & 52.66 & 60.30& 70.54& 37.26& 68.02 &  58.73 & 45.98&52.14&76.15 \\
TRAR~\cite{trar}      & ICCV'2021   & MUL  & 69.28 & 62.46&  70.96& 81.54 & 59.30& 69.23 & 78.87& 46.63& 56.56&82.49  \\
RSAdapter~\cite{RSAdapter}   & TRGS'2024   & MUL  & 64.90 & 58.93& 66.70& 78.66 & 39.27 & 69.20 & 59.15 & 48.46 & 57.87 & 84.25  \\
HRVQA~\cite{HRVQA}    & ISPRS'2024  & MUL  & 68.65 & 62.18 & 69.59 & 81.65& \textbf{60.44}& 68.73 & 77.89 & 45.63 & 55.45 & 82.90  \\
\textbf{Ours} & -      & MUL  & \textbf{71.89} & \textbf{65.12} & \textbf{73.27}& \textbf{85.30}& 59.35 & \textbf{71.04} & \uline{79.37}& 48.08 & \textbf{60.41} &  \textbf{87.81}  \\

\hline
\end{tabular}
\label{table3}
\begin{tablenotes}
        \footnotesize
        \item[]
        \begin{center}
            $\textbf{Best}$, $\uline{Second\enspace best}$. Higher metrics are better
        \end{center}
      \end{tablenotes}
\end{threeparttable}
\end{table*}

\textbf{Quantitative Evaluation.}
Table\ref{table3} presents the comparative results of the aforementioned methods on the proposed dataset. To highlight the performance of different methods across specific question types, we separately present the results of the models on four major and four minor question types.

The comparison results demonstrate that our model, TGFNet, significantly outperforms all other models. Specifically, TGFNet achieves the highest AA at 71.89\% and OA at 65.12\%, reflecting improvements of 1.92\% (from 69.97\% to 71.89\%) and 0.97\% (from 64.15\% to 65.12\%), respectively. TGFNet delivers the best performance across various question categories, including "compare", "presence", "number", "shape", and "theme". Notably, for the "compare", "presence", and "number" types, TGFNet achieves significant accuracy increases of 1.97\%, 1.46\%, and 1.36\%, respectively, relative to the second-best method. However, for the "location", "deduce", and "length" question categories, TGFNet's performance is relatively average. Nonetheless, the gap compared to the highest accuracy is not substantial.

\textbf{Qualitative Evaluation.}
Our experiments reveal that single optical and SAR image inputs each exhibit distinct strengths across specific question categories.
%a single optical image input and a single SAR image input exhibit inconsistent OA across different methods. However, for specific question categories, the performance remains relatively consistent. 
SAR images tend to perform better in categories requiring spatial structure information, such as "location", "length" and "number". Conversely, optical images perform better in categories requiring detailed texture information, such as "compare", "presence" and "shape". This is likely because SAR images are less affected by cloud cover and low-light conditions, making spatial structure information (e.g., location and object boundaries) clearer. Meanwhile, optical images contain richer detail and texture information (e.g., color and contours), providing more distinctive semantic information. This suggests that both optical and SAR images have their respective advantages.

Additionally, after fusing optical and SAR images, the overall performance and performance on specific question categories for most methods are superior to either single optical image input or single SAR image input, often outperforming both. For example, Bi-Modal~\cite{bazi2022bi}, after fusing optical and SAR images, improves OA and AA by 0.97\% and 1.16\%, respectively, compared to the best-performing single input modality model. It also achieves improvements of 0.26\%, 1.43\%, 0.47\%, 2.12\%, and 1.03\% in the "compare", "presence", "number", "deduce" and "shape" question categories, respectively, relative to the second-best modality. This indicates that the complementary information from optical and SAR images can be effectively fused, enhancing the model's robustness in VQA tasks.

Despite the fusion of optical and SAR images, Bi-Modal~\cite{bazi2022bi} fails to achieve the highest accuracy in the "location", "length" and "theme" question categories. This may be due to the simple addition operation failing to effectively achieve the complementary fusion of optical and SAR images. It also suggests that due to the limitations of the fusion network, the fused features are not always effective. Therefore, it is crucial to perform a more comprehensive integration of complementary information from optical images, SAR images, and fused images, ultimately improving the model’s robustness.

In summary, our experiments show that fusing optical and SAR images generally improves performance compared to single-modality inputs. However, the effectiveness of fused features can be limited by the fusion methods and model architectures. Our proposed method, TGFNet, addresses this by focusing on relevant image content through a coarse-to-fine approach and adaptively integrating complementary information from optical, SAR, and fused images. This allows TGFNet to fully leverage the strengths of different modalities, achieving superior performance on the proposed dataset and surpassing SOTA methods in both AA and OA.

\begin{table}[h]\footnotesize
\centering
\normalsize
\caption{Ablation study of AMEF, CFAR, and RQAF. RQAF stands for Regional Quality-Aware Fusion Network, AMEF stands for the Adaptive Multi-Expert Fusion Module without RQAF, and CFAR stands for the Text-guided Coarse-to-Fine Attention Refinement Module.}
\renewcommand\arraystretch{1.2}
\tabcolsep=0.5cm
\resizebox{\linewidth}{!}{
\begin{tabular}{cccccc}
\hline
        ~ & AMEF & CFAR & RQAF & OA & AA \\ \hline
        Exp1 & - & - & - & 70.38 & 62.45 \\ 
        Exp2 & $\surd$ & - & - & 70.72 & 64.60 \\ 
        Exp3 & $\surd$  & $\surd$ & - & 71.35 & 64.63 \\ 
        Exp4 & $\surd$ & $\surd$ & $\surd$ & \textbf{71.89} & \textbf{65.12} \\ \hline
    \end{tabular}
}
\label{table4}
\end{table}

\begin{table}[!th]\footnotesize
\centering
\setlength{\tabcolsep}{10mm}{
\normalsize
\caption{Ablation results for different fusion methods, where Add refers to pixel-level summation and Concat refers to direct connection}
\label{table5}
\renewcommand\arraystretch{1.2}
\resizebox{\linewidth}{!}{
% \resizebox{\columnwidth} {!}{

\begin{tabular}{ccc}
\hline
Integration Method  & OA & AA \\ \hline
        OPT & 68.83&60.61   \\ 
        SAR & 68.46&61.02 \\ 
        Add & 69.24&62.04 \\ 
        Concat & 69.42&62.35 \\ 
        Transformer~\cite{transformer} &69.47&63.19 \\ 
        \textbf{TGFNet} &\textbf{71.89} & \textbf{65.12} \\ \hline
\end{tabular}
}
}
\end{table}

\subsection{Ablation Experiments}
\label{Ablations}

To validate the effectiveness of the proposed modules and fusion methods, we conduct several ablation studies.\par

\textbf{Effectiveness of Proposed Modules.} The proposed TGFNet consists of two main modules: the Text-guided Coarse-to-Fine Attention Refinement (CFAR) module and the Adaptive Multi-Expert Fusion (AMEF) module. The AMEF utilizes a two-stage fusion strategy to achieve a complementary fusion of optical and SAR images. In the first stage, the Regional Quality-Aware Fusion (RQAF) network performs a patch-level fusion of optical and SAR images. In the second stage, the Adaptive Fusion (AF) network adaptively integrates the prediction results from the OPT Expert, the SAR Expert, and the Fusion Expert, producing the final prediction. To validate the effectiveness of the proposed modules, including the AMEF, CFAR, and RQAF within the AMEF, we designed a series of TGFNet variants. Each variant is defined as follows: 

\begin{itemize}   

 \item \textbf{Exp1:} In this experiment, both modules of TGFNet are removed, and the fusion of optical and SAR images is carried out using a simple addition operation.       

 \item \textbf{Exp2:} This experiment retains only the AMEF, replacing the RQAF module with a simple addition operation.        
 \item \textbf{Exp3:} Building on Exp2, the CFAR is introduced. This module is designed to identify regions relevant to the query within complex remote sensing scenes.        

\item \textbf{Exp4:} Building on Exp3, the RQAF is integrated, finalizing the TGFNet architecture. This module primarily enhances patch-level complementary fusion of different modality images by accounting for the quality of each imaging modality.

\end{itemize}

Table\ref{table4} presents the comparative performance of various TGFNet variants on the proposed dataset. Notably, the complete TGFNet (Exp4) surpasses all other variants, while the baseline model (Exp1) registers the lowest performance. Specifically, Exp2 achieves improvements of 0.34\% in OA and 2.15\% in AA over Exp1. This suggests that the proposed AMEF module can effectively facilitate the complementary fusion of optical and SAR images, even in the absence of the RQAF. Exp3 demonstrates further enhancements in OA and AA by 0.63\% and 0.03\%, respectively, compared to Exp2. This underscores the beneficial impact of incorporating the CFAR, which contributes significantly to overall model performance. Exp4 exhibits an additional increase of 0.54\% in OA and 0.49\% in AA relative to Exp3. This result highlights the significant performance gains realized by integrating the RQAF into the AMEF, thereby substantiating the effectiveness of the RQAF. Finally, all variant approaches demonstrate superior performance relative to the baseline (Exp1), underscoring the successful integration and utility of the proposed modules.

\textbf{Effectiveness of Fusion Methods.} To assess the effectiveness of the proposed fusion methods, we conducted a comparative analysis of different fusion strategies: add-based, concat-based, and transformer-based. As illustrated in Table\ref{table5}, TGFNet consistently delivers superior performance across all fusion approaches, underscoring the efficacy of our fusion method. It is evident that VQA models employing optical-SAR fusion methods consistently outperform those relying on a single image modality input, particularly in terms of OA and AA. This suggests that the limitations of individual image modalities can hinder the performance of VQA models.
%This observation suggests that the inherent limitations of individual image modalities can significantly impede the performance of VQA models. 
When evaluating the alternative fusion methods, it is clear that models utilizing the add-based fusion method achieve the least favorable results, while those employing the concat-based method show moderate improvement. Notably, models incorporating the transformer-based fusion method demonstrate the most significant enhancements in both OA and AA. This improvement is likely attributable to the cross-attention mechanism's capability to effectively facilitate cross-modal feature fusion.\par

\section{Conclusion}
In this paper, we address critical challenges in RSVQA under adverse imaging conditions through several key contributions. We introduce OSVQA, the first large-scale optical-SAR alignment dataset for RSVQA, comprising 6,008 image pairs and 1,036,694 question-answer pairs across diverse categories.
%comprising 6008 image pairs and  over 1,000,000 question-answer pairs across diverse types. 
This dataset serves as a comprehensive benchmark for evaluating RSVQA methods under challenging conditions. We propose TGFNet, a novel text-guided optical-SAR fusion network that effectively leverages complementary information from both modalities. TGFNet incorporates a Text-guided Coarse-to-Fine Attention Refinement (CFAR) module, and an Adaptive Multi-Expert Fusion (AMEF) module to enhance performance in complex scenarios. Extensive experiments on OSVQA demonstrate that TGFNet significantly outperforms existing methods, achieving SOTA results in both Average Accuracy and Overall Accuracy. These advancements highlight the potential of multi-modal fusion in enhancing RSVQA robustness for real-world applications. Future work will focus on exploring unsupervised annotation methods, investigating advanced fusion techniques, and developing open-ended, multi-source RSVQA systems.

\section*{Declaration of Competing Interest}
The authors declare that they have no known competing financial interests or personal relationships that could have appeared to influence the work reported in this paper.

\section*{Acknowledgement}
This work was supported in part by the National Natural Science Foundation of China(No. 62306005, 62006002, and 62076003), in part by the Joint Funds of the National Natural Science Foundation of China(No. U20B2068), in part by the Natural Science Foundation of Anhui Province(No. 2208085J18 and 2208085QF192), and in part by the Natural Science Foundation of Anhui Higher Education Institution (No. 2022AH040014).

\bibliographystyle{unsrt}

\bibliography{mycankaowenxian}

\end{sloppypar}
\end{document}